\documentclass{bmvc2k}
\usepackage{epsfig}
\usepackage{graphicx}
\usepackage{amsmath}
\usepackage{amssymb}
\usepackage{algorithm}
\usepackage[noend]{algpseudocode}
\usepackage{pdfpages} 
\usepackage{booktabs}
\usepackage{enumitem}
\usepackage{layouts}

\usepackage{color}
\usepackage[draft,inline,nomargin,noindex]{fixme}

\title{Reciprocal Attention Fusion for Visual Question Answering}

\addauthor{Moshiur R Farazi$^{\text{1,}}$}{moshiur.farazi@anu.edu.au}{2}
\addauthor{Salman H Khan$^{\text{2,1,}}$}{salman.khan@anu.edu.au}{3}

\addinstitution{
Australian National University, \\
 Australia
}
\addinstitution{
 Data61 - CSIRO, Australia
}
\addinstitution{
 Inception Institute of AI, UAE
}

\runninghead{Farazi, Khan}{Reciprocal Attention Fusion for VQA}

\def\etal{\emph{et al}\bmvaOneDot}

\begin{document}

\maketitle

\begin{abstract}
Existing attention mechanisms either attend to local image-grid or object level features for Visual Question Answering (VQA). Motivated by the observation that questions can relate to both object instances and their parts, we propose a novel attention mechanism that jointly considers reciprocal relationships between the two levels of visual details. The bottom-up attention thus generated is further coalesced with the top-down information to only focus on the scene elements that are most relevant to a given question. Our design hierarchically fuses multi-modal information i.e., language, object- and grid-level features, through an efficient tensor decomposition scheme. The proposed model improves the state-of-the-art single model performances from 67.9\% to 68.2\% on VQAv1 and from 65.7\% to 67.4\% on VQAv2, demonstrating a significant boost. 
\end{abstract}

\section{Introduction}
\label{sec:intro}

An AI agent equipped with visual question answering ability can respond to intelligent questions about a complex scene. This task bridges the gap between visual and language understanding to realize the longstanding goal of highly intelligent machine vision systems. Recent advances in automatic feature learning with deep neural networks allow joint processing of both visual and language modalities in a unified framework, leading to significant improvements on the challenging VQA problem \cite{antol2015vqa,krishna2016visual,johnson2016clevr,zhu2016visual7w,goyal2016making}.

To deduce the correct answer, an AI agent needs to correlate image and question information.  A predominant focus in the existing efforts has remained on attending to local regions on the image-grid based on language input \cite{xu2015show, lu2016hierarchical, yang2016stacked, jabri2016revisiting, shih2016look}. Since these regions do not necessarily correspond to representative scene elements (objects, attributes and actions), there exists a "semantic gap" in such attention mechanisms. To address this issue, Anderson \etal \cite{Anderson2017up-down} proposed to work at the object level, where model attention is spread over a set of possible object locations. However, the object proposal set considered in this way is non-exhaustive and can miss important aspects of a scene. Furthermore, language questions can pertain to local details about objects parts and attributes, which are not encompassed by the object-level scene decomposition. 

\begin{figure*}
\begin{center}
\includegraphics[width=.9\linewidth]{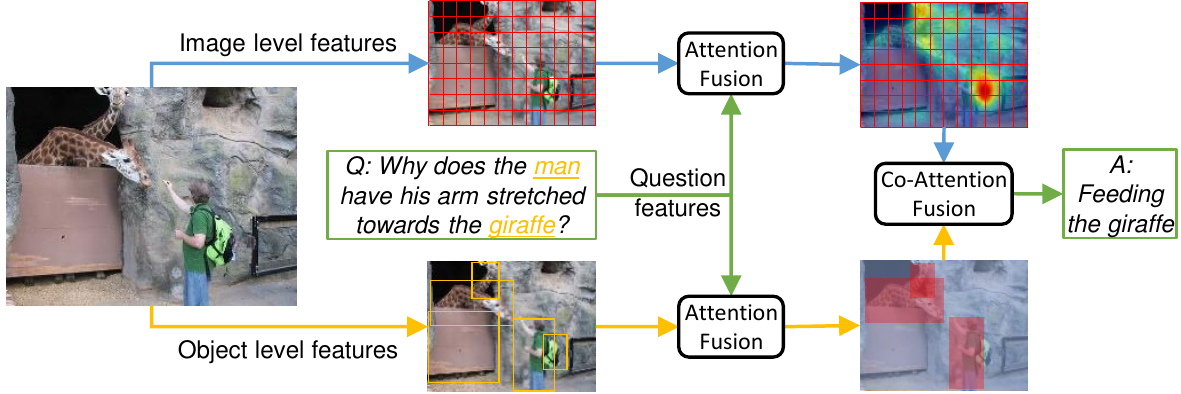}
\end{center}
  \vspace{-1.5em}
\caption{Applying attention to reciprocal visual features allow a VQA model to obtain the most relevant informations required to answer a given visual question.}
\label{fig:1}
\end{figure*}
  \vspace{0.5em}
In this work, we propose to simultaneously attend to both low-level visual concepts as well as the high-level object based scene representation. Our intuition is based on the fact that the questions can be related to objects, object-parts and local attributes, therefore focusing on a single scene representation can degrade model capacity. To this end, we jointly attend to two reciprocal scene representations that encompass local information on the image-grid and the object-level features. The bottom-up attention thus generated is further combined with the top-down attention driven by the linguistic input. Our design draws inspiration from the human cognitive psychology, where attention mechanism is known to be a combination of both exogenous (bottom-up) and endogenous (top-down) factors \cite{desimone1995neural, borji2013state}. 

Given the multi-modal inputs, a critical requirement is to effectively model complex interactions between the multi-level bottom-up and top-down factors. For this purpose, we propose a multi-branch CNN architecture that hierarchically fuses visual and linguistic features by leveraging an efficient tensor decomposition mechanism \cite{tucker1966some, benyounescadene2017mutan}. Our experiments and extensive ablative study proves that a language driven attention on both image-grid and object level representation  allows a deep network to model the complex interaction between vision and language as our model outperforms the state-of-the-art models in VQA tasks. 

In summary, this paper makes the following key contributions:
\begin{itemize}[topsep=0pt,itemsep=-1ex,partopsep=1ex,parsep=1ex]
\item A hierarchical architecture incorporating both the bottom-up and top-down factors pertaining to meaningful scene elements and their parts. 
\item Co-attention mechanism enhancing scene understanding by combining local image-grid and object-level visual cues.
\item Extensive evaluation and ablation on both balanced and imbalanced versions of the large-scale VQA dataset achieving single model state-of-the-art performance in both.
\end{itemize}

\section{Related Works}
\textbf{Deep Networks:}  
Given the success of deep learning, one common approach to address the VQA problem is by generating image features using pretrained Convolutional Neural Networks (CNNs), e.g., VGGNet \cite{simonyan2014very}, ResNet \cite{he2016deep}, and language features using word-embeddings or Long Short-Term Memory (LSTM) \cite{antol2015vqa}. After generating image and language features, some approaches train RNNs to generate top-K candidate answers and use a multi-class classifier to choose the best answer \cite{antol2015vqa,zhu2016visual7w,zhou2015simple}. A number of attention mechanisms have been incorporated within deep networks to automatically focus on specific details in an image based on the given question \cite{lu2016hierarchical,yang2016stacked,jabri2016revisiting}.  Memory networks have also been incorporated in many top performing models \cite{xiong2016dynamic, sukhbaatar2015end, zhuknowledge} where the questions required the system to compare attributes or use a long reasoning chain. While robust features and memory modules help capture some aspects of the semantics present in the scene, modeling the complex interplay between image-grid and objects level features can complement the understanding of the rich scene semantics.


\textbf{Attention Models:}
The incorporation of spatial attention on the image and/or the text features has been investigated to capture the most important parts required to answer a given question \cite{xu2015show,yang2016stacked,jabri2016revisiting,shih2016look,lu2016hierarchical}. 
Different pooling methods have been used previously to compute the attention maps such as soft attention, bilinear pooling and tucker fusion \cite{xu2015show,gao2016compact,benyounescadene2017mutan}. All these techniques explore top-down attention and only focus on the image-grid. 
Different to these works, based on the observation that questions pertain to objects, their parts and attributes, we propose to work jointly at the spatial grid of image regions and the object-level. The closest to our work is \cite{Anderson2017up-down}, which attends to salient objects in an image for improved  VQA. However, they ignore two key aspects of visual reasoning i.e., the image level visual features and an effective fusion mechanism to combine the bimodal interaction between visual and language features. Another recent effort \cite{lu2017co} co-attends to both image regions and objects, but uses a simplistic fusion mechanism based on element-wise multiplication that is outperformed by our bilinear feature encoding. Besides, our multi-level attention mechanism effectively uses object features  and scene context based on the natural language queries.


\section{Methods}
The VQA task requires an AI agent to generate a natural language response, given a visual (i.e. image, video) and natural language input (i.e.\ questions, parse). We formulate VQA task as a classification task, where the model predicts the correct answer ($\hat{a}$) from all possible answers for a given image (\textbf{v}) and question (\textbf{q}) pair:
\begin{equation}
\hat{a} = \operatorname*{argmax}_{a \in A} p(a|\mathbf{v},\mathbf{q}; \theta),
\end{equation}
where $\theta$ denotes the set of parameters used to predict the best answer from the set of all possible answers $A$. 

\begin{figure*}[!ht]
\begin{center}
\includegraphics[width=\linewidth]{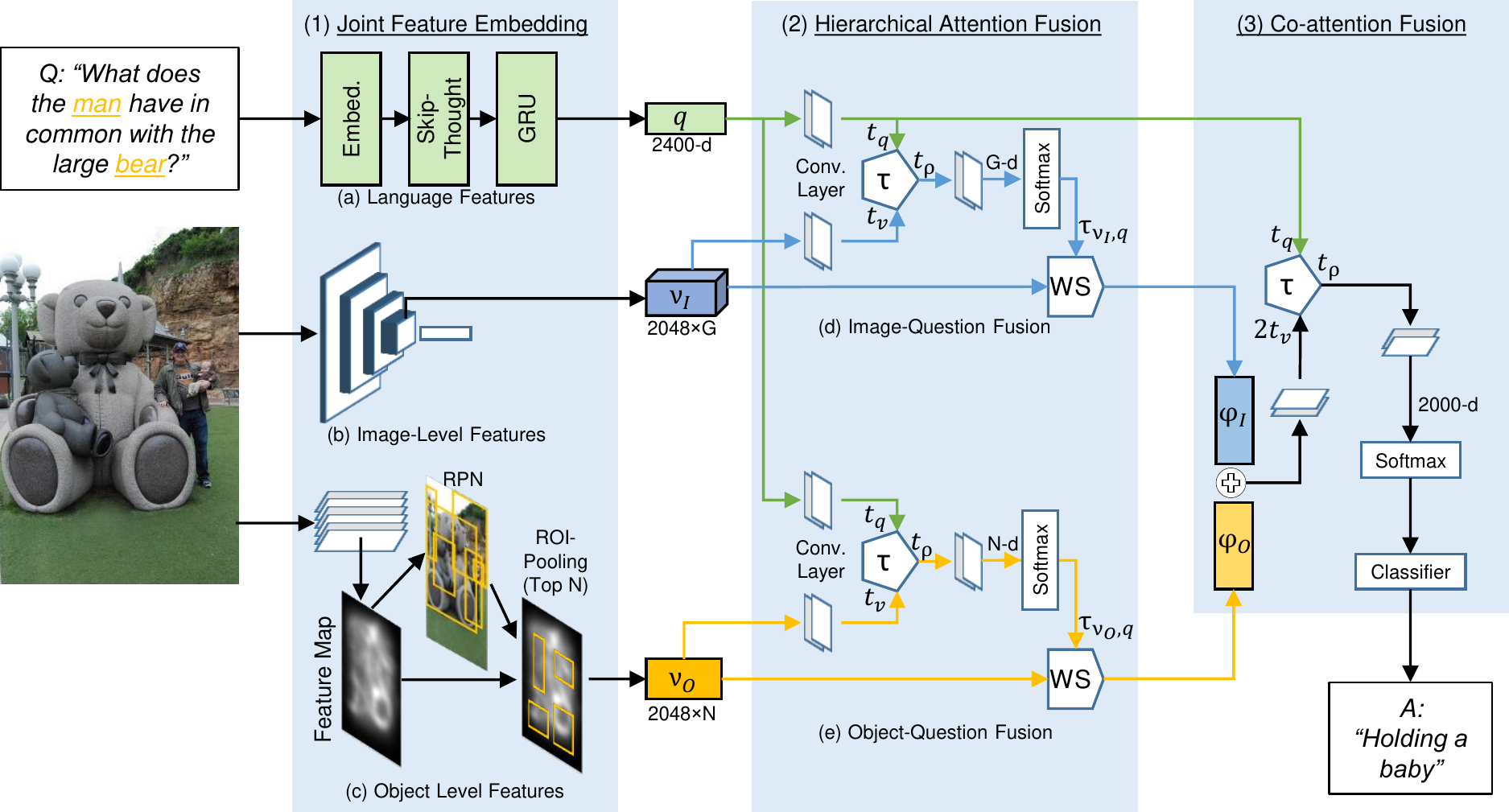}
\end{center}
 \vspace{-1.5em}
\caption{Given an image-question pair, our model employs (1) Joint Feature Embedding (Sec.\ref{sec:featureEmbedding}) to embed (a) Language Feature $q$, (b) Image-Level Feature $v_I$ and (c) Object-Level Feature $v_O$. Further, these embeddings undergo (2) Hierarchical Attention Fusion (Sec.\ref{Sec:HieAttFusion}) which consists of (d) Image-Question and (e) Object-Question Fusion followed by top-down attention. These multi-modal representations are combined together by (3) Co-attention Fusion (Sec.\ref{sec:coattnfusion}) that predicts an answer for the given Image-Question pair. Overall, the proposed model attends to complementary levels of scene details and fuses multi-modal information to predict highly accurate answers.}
\label{fig:2} 
\end{figure*}

Our proposed architecture to perform VQA task is illustrated in Figure \ref{fig:2}. The key highlights of our proposed architecture include a hierarchical attention mechanism that focuses on complementary levels of scene details i.e., grid of image regions and object proposals. The relevant co-attended features are then fused together to perform final prediction. We name our model as the \emph{`Reciprocal Attention Fusion'} because it simultaneously attends to two complementary scene representations i.e., image-grid and object proposals. Our experimental results demonstrate that both levels of scene details are reciprocal and reinforce each other to achieve the best single-model performance on challenging VQA task. Before elaborating on the hierarchical attention and feature fusion, we first discuss the joint feature embedding in Section \ref{sec:featureEmbedding}.

\subsection{Joint Feature Embedding}
\label{sec:featureEmbedding}
Let $ V $ be the collection of all visual features extracted from an image and $ Q $  be the language features extracted from the question. The objective of joint embedding is to learn the language feature representation $ q = \chi (Q) $ and multilevel visual features $ v_k = \zeta (V) $. These feature representations are used to encode the multilevel relationships between question and image which in turn is used to train the classifier to select the correct answer. 

\textbf{Multilevel visual features:} The multilevel visual embedding $v_k$ consists of image level features $v_I$ and object level features $v_O$. Our model employs ResNeXt \cite{Xie2016} to obtain image level features, $v_I \in \mathbb{R}^{n_v \times G}$ by taking the output of convolution layer before the final pooling layer, where $G$ denotes the number of spatial grid locations of the extracted visual feature with $n_v$ dimensions. This convolution layer retains the spatial information of the original image and enable the model to apply attention on the image-grid. On the other hand, our model employs object detectors to localize object instances and pass them through another deep CNN to generate object level features $v_O \in \mathbb{R}^{n_v \times N}$ for $N$ object proposals. We use Faster R-CNN \cite{ren2015faster} with ResNet-101 \cite{he2016deep} backbone and pretrain the object detector on ImageNet \cite{deng2009imagenet} and again retrain it on Visual Genome Dataset \cite{krishna2016visual} with class label and attribute features similar to \cite{Anderson2017up-down}.

\textbf{Bottom-up (BU) Attention:} In order to focus on the most relevant features, two bottom-up attention mechanisms are applied during multilevel feature extraction. The image-grid attention is generated using ResNeXt \cite{Xie2016} pretrained on ImageNet \cite{deng2009imagenet} to obtain $v_I \in \mathbb{R}^{2048 \times 14 \times 14}$, which represents $ 2048 $ dimensional features vectors for $G = 14 \times 14$ image-grid over the visual input. The size and scale of the image-grid can be changed by using different CNN architecture or taking the output of a different convolutional layer to generate a different sized BU attention. Meanwhile, object proposals are generated in a bottom up fashion to encode object level visual features $v_O$. We select a total of top $N =36$ object proposals whose $n_v = 2048$ dimensional feature vectors are obtained from the ROI pooling layer in the Region Proposal Network. 

\textbf{Language features:} To represent the questions embedding in an end-to-end framework, GRUs \cite{cho2014properties} are used in a manner similar to \cite{fukui2016multimodal,benyounescadene2017mutan}. The words in questions are encoded using one-hot-vector representation and embedded into vector space by using  a word embedding matrix. The embedded word vectors are fed to the GRU with $n_q$ units initialized with pretrained Skip-thought Vector model \cite{kiros2015skip}. The output of the GRU is fine-tuned to get the language feature embedding $q = \{q_i \; : \, i \in [1,n_q]\}$ where $n_q = 2400$. The language feature embedding is used to further refine the spatial visual features (i.e. image-grid and object level) by incorporating top-down attention discussed in Section \ref{Sec:HieAttFusion}.

\subsection{Hierarchical Attention Fusion}
\label{Sec:HieAttFusion}
The hierarchical attention mechanism takes spatial visual features $v_I,v_O$ and language feature $q$ as input and a learns multi-modal representation $W$ to predict answer embedding  $\rho$. This step can be formulated as an outer product of the multi-modal representation, visual and language embeddings as follows:
\begin{equation}
\label{eq:1}
\rho = W \times_{1} \ q \ \times_{2} \ v,
\end{equation}
where, $\times_{n}$ denotes n-mode tensor-matrix product. However, this approach has some serious practical limitations in terms of learnable parameters for $W \in \mathbb{R}^{n_v \times n_q \times n_{\rho}}$ as the visual and language feature are very high dimensional, which results in huge computational and memory requirements. To counter this problem, our model employs a multi-modal fusion operation $\tau(v,q)$ to encode the relationships between these two modalities, which is discussed next.

\textbf{Multi-modal Fusion:} 
Multi-modal fusion aims to reduce the number of free parameters in tensor $W \in \mathbb{R}^{n_v \times n_q \times n_{\rho}}$ for a fully parameterized VQA bilinear model. Our model achieves this by using Tucker Decomposition \cite{tucker1966some} which is a special case of higher-order principal component analysis to express $W$ as a core tensor $T_c$ multiplied by a matrix along input mode. The decomposed tensors are fused in a manner similar to \cite{benyounescadene2017mutan} that encompass the multi-modal relationship between language and vision domain. The tensor $W$ can be approximated as:
\begin{equation}
W \approx T_c \times_{1} T_q \times_{2} T_v \times_{3} T_{\rho} = [\![ T_c ; T_q, T_v, T_{\rho} ]\!]
\end{equation}
where $T_v \in \mathbb{R}^{n_v \times t_v}$, $T_q \in \mathbb{R}^{n_q \times t_q}$ and $T_{\rho} \in \mathbb{R}^{n_{\rho} \times t_{\rho}}$ are factor matrices similar to principal components along each input and output embeddings and $T_c \in \mathbb{R}^{t_v \times t_q \times t_{\rho}}$ is the core tensor which encapsulates interactions between the factor matrices. The notation $[\![ \cdot ]\!]$ represents the shorthand for Tucker decomposition. In practice, the decomposed version of $W$ is significantly smaller number of parameters than the original tensor \cite{bader2007efficient}. 

After reducing the parameter complexity of $W$ with tucker decomposition, the fully parametrized outer product representation in Eq. \ref{eq:1} can be rewritten as:
\begin{equation}
\rho = T_c \ \times_{1} \ \tilde{q} \ \times_{2} \  \tilde{v} \ \times_{3} \ T_{\rho}
\end{equation}
where $\tilde{v} = v^{\intercal} \ T_v  \in \mathbb{R}^{t_v}$ and $\tilde{q} = q^{\intercal} \ T_q  \in \mathbb{R}^{t_q}$. We define a prediction space $\rho = \tau^{\intercal} \ T_{\rho} \in \mathbb{R}^{n_{\rho}}$ where the multi-modal fusion $\tau$ is:
\begin{equation}
\label{eq:tucker_fusion}
\tau = T_c \ \times_{1} \ \tilde{q} \ \times_{2} \ \tilde{v} \ \in \mathbb{R}^{t_{\rho}} 
\end{equation}

The Tucker decomposition allows our model to decompose $W$ into a core tensor $T_c$ and three matrices. The first two matrices, $T_q$ and $T_v$ project the question and visual embeddings to lower $t_q$ and $t_v$ dimensional space that learns to model the multi-modal interaction and projects the resulting output to $t_{\rho}$ dimensional vector. We set the input projections dimension to $t_q = t_v = 310$ and output projection dimension as $t_{\rho} = 510$. The input and output tensor projection dimensions determine the complexity of the model and the degree of multi-modal interaction which in turn affects the performance of the model. These values are set empirically by testing them on VQAv1 validation dataset. It has been reported in the literature \citep{fukui2016multimodal,benyounescadene2017mutan} that applying nonlinearity to the input feature embeddings improve performance of multi-modal fusion. Therefore, we encode $\tilde{v}$ and $\tilde{q}$ with $\tanh$ nonlinearity during fusion. The output of the multi-modal fusion $\tau \in \mathbb{R}^{t_{\rho}}$ passes through convolution and softmax layers to create $1 \times G$ and $1 \times N$ dimensional representation for image-question and object-question embedding respectively. Thus, by employing hierarchical attention fusion, we embed question with spatial visual features to generate image-question $\tau_{v_I,q} \in \mathbb{R}^{1 \times G}$ and object-question $\tau_{v_O,q} \in \mathbb{R}^{1 \times N}$ embedding. 

\textbf{Top-down (TD) Attention} The image level and object level features are used alongside image-question and object-question embeddings to generate an attention distribution over spatial grid and object proposals respectively. We take weighted sum (WS) of the spatial visual features (i.e.\ $v_I$ and $v_O$) vectors using the attention weights (i.e.\ $\tau_{v_I,q}$ and $\tau_{v_O,q}$) to generate $\varphi_I$ and $\varphi_O$ which are top-down attended visual features,
\begin{equation}
\label{eq:weightedsum}
\varphi_I = \sum_{i}^{G} \tau_{v_I,q}^i \ v_I^i \quad \textrm{and} \quad \varphi_O = \sum_{i}^{N} \tau_{v_O,q}^i \ v_O^i.
\end{equation} 

\subsection{Co-attention Fusion}
\label{sec:coattnfusion}
The attended image-question and object-question visual features represent a combination of visual and language features that are most important to generate an answer for a given question. We concatenate these two bimodal representations to create the final visual-question embedding $\varphi = \varphi_I \oplus \varphi_O$. The visual-question embedding, $\varphi$ and original question embedding $q$ again undergo same multi-modal fusion as Eq. \ref{eq:tucker_fusion}. The only difference is now $t_{\varphi} = 2 \times t_v $ as our model uses two glimpse attention which was found to yield better results \cite{fukui2016multimodal, benyounescadene2017mutan,kim2016hadamard}. The output of the final fusion is then passed on to the classifier that predicts the best answer $\hat{a}$ from the answer dictionary $A$ given question $ \bf{q} $ and visual input $ \bf{v}$. 

\begin{table*}[t]
  \centering
  \scalebox{0.9}{
  \begin{tabular}{l|cccc|cccc}
    \toprule
    Methods									& \multicolumn{4}{c}{Test-dev}	& \multicolumn{4}{|c}{Test-standard}	\\			
    									& Y/N 	& No. 	& Other & All 	& Y/N 	& No. 	& Other & All	\\ 
    \midrule																				
    RAF (Ours)							& \textbf{85.9}	& \textbf{41.3}	& \textbf{58.7} 	& \textbf{68.0}	
    									& \textbf{85.8}	& 41.4	& 58.9 	& \textbf{68.2}		\\
    \midrule
    ReasonNet\cite{ilievski2017multimodal}
    									& -	  	& -	  	& -	 	& -	 	& 84.0 	& 38.7 	& \textbf{60.4}	& 67.9	\\
    MFB+CoAtt+Glove \cite{yu2018beyond}	& 85.0  & 39.7  & 57.4  & 66.8  & 85.0 	& 39.5 	& 57.4 	& 66.9	\\
    Dual-MFA \cite{lu2017co}			& 83.6  & 40.2  & 56.8  & 66.0  & 83.4 	& 40.4 	& 56.9 	& 66.1	\\ 
    MLB+VG \cite{kim2016hadamard}		& 84.1 	& 38.0  & 54.9  & 65.8  & -	 	& -	 	& -	 	& -		\\	
    MCB+Att+GloVe \cite{fukui2016multimodal}
    									& 82.3  & 37.2  & 57.4  & 65.4  & -		& -		& -		& -		\\
    MLAN \cite{yu2017multi}				& 81.8  & 41.2  & 56.7  & 65.3  & 81.3 	& \textbf{41.9} & 56.5 	& 65.2	\\
    MUTAN \cite{benyounescadene2017mutan}\footnotemark
    									& 84.8  & 37.7  & 54.9  & 65.2  & - 	& -  	& - 	& -		\\
    DAN (ResNet) \cite{nam2016dual}		& 83.0  & 39.1  & 53.9  & 64.3  & 82.8 	& 38.1 	& 54.0 	& 64.2	\\
    HieCoAtt \cite{lu2016hierarchical}	& 79.7	& 38.7	& 51.7	& 61.8	& -		& -		& -		& 62.1 	\\
    A+C+K+LSTM\cite{wu2016ask}			& 81.0  & 38.4  & 45.2  & 59.2  & 81.1 	& 37.1 	& 45.8 	& 59.4	\\
    VQA LSTM Q+I \cite{antol2015vqa}	& 80.5	& 36.8	& 43.1	& 57.8	& 80.6	& 36.5	& 43.7	& 58.2	\\
    SAN\cite{yang2016stacked}			& 79.3	& 36.6	& 46.1	& 58.7	& -		& -		& -		& 58.9	\\
    AYN \cite{malinowski2017ask}			& 78.4  & 36.4  & 46.3  & 58.4  & 78.2 	& 37.1 	& 45.8 	& 59.4	\\
    NMN \cite{andreas2016neural}			& 81.2	& 38.0	& 44.0	& 58.6	& -		& -		& -		& 58.7	\\
    DMN+ \cite{xiong2016dynamic}			& 60.3	& 80.5	& 48.3	& 56.8	& -		& -		& -		& 60.4	\\
    iBowling \cite{zhou2015simple}		& 76.5	& 35.0	& 42.6	& 55.7	& 76.8	& 35.0	& 42.6	& 55.9	\\    
    \bottomrule
  \end{tabular}}
  \vspace{0.5em}
  \caption{Comparison of the state-of-the-art methods with our single model performance on VQAv1.0 test-dev and test-standard server.}
  \label{tab:1}
\end{table*}


\section{Experiments}
\subsection{Dataset}
We perform experiments on {\bf VQAv1} \cite{antol2015vqa} and {\bf VQAv2} \cite{goyal2016making} both of which are large scale VQA datasets. VQAv1 contains over 200K images from the COCO dataset with 610K natural language open-ended questions. VQAv2 \cite{goyal2016making} contains almost twice as many question for the same number of images. VQAv2 has a balanced image-question pair to mitigate the language bias that allows a more realistic evaluation protocol. \textbf{Visual Genome} is another larger scale dataset that has image question pair with dense annotation of objects, attributes \cite{krishna2016visual}. We train a pretrained faster RCNN model (on ImageNet) again on Visual Genome dataset with class and attribute labels to extract object level features from the input image.

\subsection{VQA Model Architecture}

\textbf{Question Feature Embedding:} Our model embeds the question features by first generating the questions and answer dictionary from training and validation set of the VQA datasets. We make the question and answers lower case, remove punctuation and perform other standard preprocessing steps before tokenizing the words, and representing them into one-hot vector representation. As mentioned in Section \ref{sec:featureEmbedding}, these question embeddings are fed to GRUs pretrained with Skip-thoughts \cite{kiros2015skip} model that generates $2400$-d language feature embeddings for the given question. When experimenting with VQAv1 and VQAv2, we parse questions respectively from training and validation sets to create the question vocabulary.

\footnotetext{Single model performance is evaluated using their publicly available code.}

\textbf{Answer Encoding:} We formulate the VQA task as a classification task. We create an answer dictionary from the training data and select the top $2000$ answers as the different classes. We pass the output of the final fusion layer through a convolutional layer that outputs a $2000$d vector. This vector is passed through the classifier to predict $\hat{a}$.

We use Adam solver \cite{kingma2014adam} with base learning rate of $10^{-4}$ and batch size of $512$ for our experiments. We keep the training parameters same for all our experiments. We use NVidia Tesla P100 (SXM2) GPUs to train our models and report our experimentation results on VQAv1 \cite{antol2015vqa} and VQAv2 \cite{goyal2016making} dataset representing $1500$ GPU hours of computation.

\begin{table*}[t]
  \centering
  \scalebox{0.9}{
  \begin{tabular}{l|cccc|cccc}
    \toprule
    Methods								& \multicolumn{4}{c}{Test-dev}	& \multicolumn{4}{|c}{Test-standard}	\\			
    									& Y/N 	& No. 	& Other & All 	& Y/N 	& No. 	& Other & All	\\ 
    \midrule																				
    RAF (Ours)							& \textbf{84.1}	& \textbf{44.9}	& \textbf{57.8} 	& \textbf{67.2}	
    									& \textbf{84.2}	& \textbf{44.4}	& \textbf{58.0} 	& \textbf{67.4}		\\
    \midrule
    BU, adaptive K \cite{teney2017tips}	& 81.8	& 44.2	& 56.1	& 65.3	& 82.2	& 43.9 	& 56.3	& 65.7		\\
    MFB \cite{yu2018beyond} 			& -		& -		& -		& 64.9	& -		& -	 	& -		& -			\\                                   
    ResonNet\cite{ilievski2017multimodal}
    									& -		& -		& -		& -		& 78.9	& 42.0	& 57.4	& 64.6		\\
   	MUTAN\cite{benyounescadene2017mutan}\footnotemark
                                        & 80.7	& 39.4	& 53.7	& 63.2	& 80.9	& 38.6 	& 54.0	& 63.5		\\
    MCB \cite{fukui2016multimodal, goyal2016making}
    									& -		& -		& - 	& -		& 77.4	& 36.7 & 51.2 	& 59.1		\\
    HieCoAtt \cite{lu2016hierarchical, goyal2016making}
    									& -		& -		& -		& -		& 71.8	& 36.5	& 46.3	& 54.6		\\
    Language only\cite{goyal2016making}	& -		& -		& -		& -	    & 67.1	& 31.6	& 27.4	& 44.3		\\
    Common answer\cite{goyal2016making} & - 	& -		& -		& -	    & 61.2	& 0.4	& 1.8	& 26.0 		\\
    \bottomrule
  \end{tabular}}
  \vspace{0.5em}
  \caption{Comparison of the state-of-the-art methods with our single model performance on VQAv2.0 test-dev and test-standard server.}
  \label{tab:2}
\end{table*}

\section{Results}
We evaluate the proposed models' performance on the VQA test servers which ensures blind evaluation on the VQAv1 \cite{antol2015vqa} and v2 \cite{goyal2016making} test sets (i.e.\ test-dev, test-standard) following the VQA benchmark evaluation approach. The accuracy $y$ of the predicted answer $\hat{a}$ is calculated with the following formulation:
\begin{equation}
y = min \bigg(\frac{\# \; \text{of humans answered $\hat{a}$}}{3}, 1 \bigg)
\end{equation}
which means that  answer provided by the model is given 100\% accuracy if at least 3 human annotators who helped create the VQA dataset gave the exact answer.

In Table \ref{tab:1}, we report VQAv1 test-dev and test-standard accuracies for our proposed RAF model and compare it with other single models found in literature. Remarkably, our model outperforms all other models in the overall accuracy. We report a  significant performance boost of $1.2\%$ on the test-dev set and $0.3\%$ on the test-standard set. It is to be noted that using multiple ensembles and data augmentation with complementary training in Visual Genome QA pairs can increase the accuracy performance of the VQA models. For instance, MCB \cite{fukui2016multimodal}, MLB \cite{kim2016hadamard}, MUTAN \cite{benyounescadene2017mutan} and MFB \cite{yu2018beyond} employ similar model ensemble consisting of 7,7,5 and 7 models respectively, and report overall $66.5$, $66.9$, $67.4$ and $69.2$ on the test-standard set. It is interesting to note that except for MFB (7) all other ensemble models are $\sim 1\%$ less than our reported single model performance. We do not ensemble our model or use data augmentation with complementary dataset as it makes the best results irreproducible and most of the models in the literature do not adopt this strategy.

\footnotetext{Performance on VQAv2 is evaluated from their publicly available repository.}

We also evaluate our model on VQAv2 test-standard dataset and compare it with state-of-the-art single model performance in Table \ref{tab:2}, illustrating that our model surpasses the closest method \cite{teney2017tips} in all question categories and overall by a significant margin of $1.7\%$. The bottom up, adaptive-k\cite{teney2017tips} is the same model whose 30-ensemble version \cite{Anderson2017up-down} reports currently the best performance among on VQAv2 test-standard dataset. This indicates our models superior capability to interpret and incorporate multi-modal relationships for visual reasoning. 

\begin{figure}[!t]
\begin{center}
\includegraphics[width=.95\linewidth]{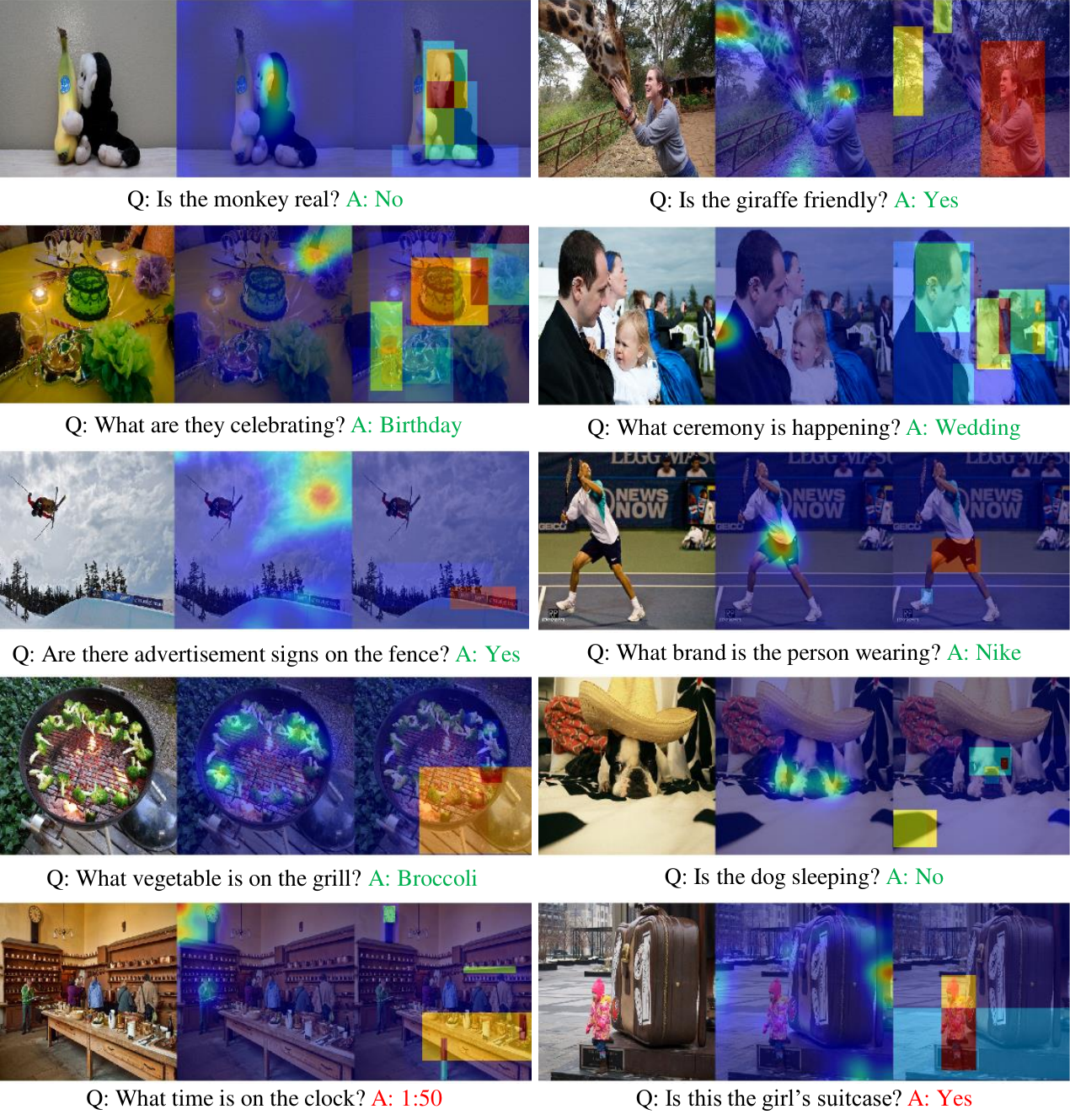}
\end{center}
  \vspace{-2.0em}
\caption{Qualitative results of the proposed Reciprocal Attention Fusion mechanism for Visual Question Answering. Given a question and an image (columns: $1,4$), attention based on image-grid (columns: $2,5$) and object proposals (columns: $3,6$) is shown above. Correct and incorrect answers are shown in \textcolor{green}{green} and \textcolor{red}{red}, respectively. Remarkably, the two attention levels provide complementary information about localized regions and objects that in turn help in obtaining the correct answer (rows: $1,2,3,4$). In some failure cases of our technique, ambiguous attention maps lead to incorrect predictions (row: $5$).}
\label{fig:4}
\end{figure}

In summary, our model achieves state-of-the-art performance on both VQAv1 and VQAv2 dataset which affirms the robustness of our model against language bias without the need of data augmentation or the use of ensemble model. We also show qualitative results in Fig.~\ref{fig:4} to demonstrate the efficacy and complimentary nature of attention focused on image-grid and object proposals.

\begin{table}[t]
\centering
    \begin{minipage}{.5\linewidth}
    \scalebox{0.9}{
        \begin{tabular}{l|l|c}
    	\toprule
    	Cat.	& Methods						& Val-set  	\\
        \midrule
    	I 	& RAF-I(ResNet)						& 53.9		\\
            & HieCoAtt \cite{lu2016hierarchical,goyal2016making} 
            									& 54.6		\\
    		& RAF-I(ResNeXt)					& 58.0		\\
        	& MCB \cite{fukui2016multimodal,goyal2016making} 
            									& 59.1		\\	
        	& MUTAN \cite{benyounescadene2017mutan} 
        										& 60.1		\\
    	\midrule
    	II	& Up-Down\cite{Anderson2017up-down}	& 63.2		\\
        	& RAF-O(ResNet)						& 63.9		\\
    	\midrule
    	III	& RAF-IO(ResNet-ResNet)				& 64.0		\\
        	& RAF-IO(ResNeXt-ResNet)		    & 64.2		\\
    	\bottomrule
    	\end{tabular}}
        \vspace{0.5em}
        \caption{Ablation Study on VQAv2 val-set.}
        \label{tab:3}
    \end{minipage} 
    \hfill \;\;
    \begin{minipage}{.45\linewidth}
		\begin{figure}[H]
              \includegraphics[width=\linewidth]{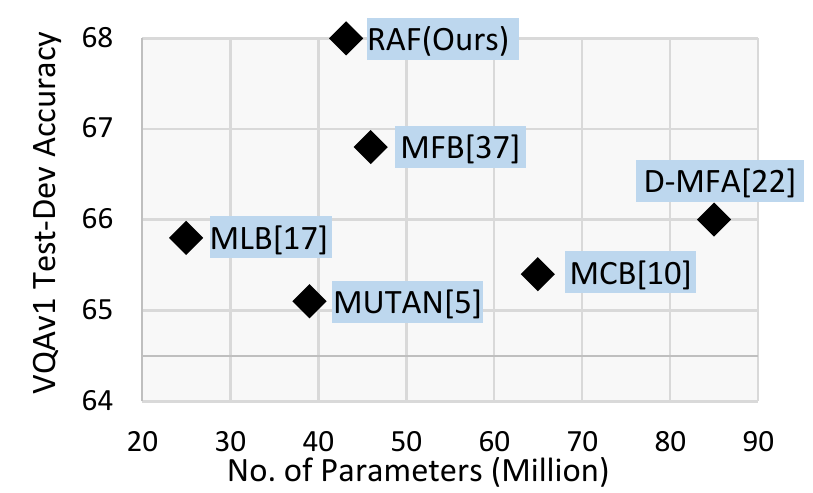}
              \vspace{-1.8em}
              \caption{Accuracy vs.\ Complexity (no. of parameters) comparison.}
              \label{fig:3}
       \end{figure}
    \end{minipage}    
\end{table}

\subsection{Ablation Study}
We perform an extensive ablation study of the proposed model on VQAv2 \cite{goyal2016making} validation dataset and compare it with the best performing model in Table \ref{tab:3}. This ablation study helps to better understand the contribution of different components of our model towards the overall performance on the VQA task. The objective of this ablation study is to show that when the language features are combined with image- grid and object level visual features, the accuracy of the high level visual reasoning task (i.e.\ VQA) increases in contrast to only combining language with image or object level features. The models reported in Category I in Table \ref{tab:3} use only image level features extracted with deep CNNs and we compare RAF-I which is a variant of our proposed RAF architecture only using image level features. We observe RAF-I achieve comparable performance in this category. In Category II, RAF-O model extracts only $36$ object level features but outperforms the models in Category I. \cite{Anderson2017up-down} also used only object level features and this variant of our model achieves comparable performance to that model. When we combine image and object level features together in Category III, we observe that the best results are obtained. This proves our hypothesis that the questions relate to both objects, object parts and local attributes, which should be attended for jointly an improved VQA performance. 

The recent Dual-MFA \cite{lu2017co} model also uses complementary image and object-level features. In contrast, our model uses more efficient bimodal attention fusion mechanism and exhibit robustness on balanced VQAv2 \cite{goyal2016making} dataset. We also study the accuracy vs.\ complexity (no. of parameters) trade off in Fig. \ref{fig:3} on VQAv1 test-dev set as most of the bilinear models do not report performance on VQAv2. Remarkably, our RAF model achieves significant performance boost over Dual-MFA (66\% to 68\%) with around half the complexity.

\section{Conclusion} 
We build our proposed model based on the hypotheses that multi-level visual features and associated attention can provide an AI agent additional information pertinent for deep visual understanding. As VQA is a standard measure of image understanding and visual reasoning, we propose a VQA model that learns to capture the bimodal feature representation from visual and language domain. To this end, we employ state of the art CNN architectures to obtain visual features for local regions on the image-grid and object proposals. Based on these feature encodings, we develop a hierarchical co-attention scheme that learns the mutual relationships between objects, object-parts and given questions to predict the best response. We validate our hypotheses by evaluating the proposed model on two large scale VQA dataset servers followed by an extensive ablation study reporting state-of-the art performance.

\section*{Acknowledgements}
The authors would like to thank CSIRO Scientific Computing team, especially Peter Campbell and Ondrej Hlinka, for their assistance in optimizing the work-flow on GPU clusters.

{\scriptsize\bibliography{egbib}}
\end{document}